# Characterizing Novelty in the Military Domain

Theresa Chadwick[1], James Chao[1], Christianne Izumigawa[1], George Galdorisi[1], Héctor Ortiz-Peña[3], Elias Loup[2], Nicholas Soultanian[1], Mitch Manzanares[1], Adrian Mai[1], Richmond Yen[1], Douglas S. Lange[1*]
1. Naval Information Warfare Center Pacific, San Diego, California
2. Naval Research Laboratory Stennis, Bay Saint Louis, Mississippi
3. CUBRC, Buffalo, New York
*- Corresponding Author: Douglas S. Lange, douglas.s.lange2.civ@us.navy.mil

## Abstract

A critical factor in utilizing agents with Artificial Intelligence (AI) is their robustness to novelty. AI agents include models that are either engineered or trained. Engineered models include knowledge of those aspects of the environment that are known and considered important by the engineers. Learned models form embeddings of aspects of the environment based on connections made through the training data. In operation, however, a rich environment is likely to present challenges not seen in training sets or accounted for in engineered models. Worse still, adversarial environments are subject to change by opponents. A program at the Defense Advanced Research Project Agency (DARPA) seeks to develop the science necessary to develop and evaluate agents that are robust to novelty. This capability will be required, before AI has the role envisioned within mission critical environments.

As part of the Science of AI and Learning for Open-world Novelty (SAIL-ON), we are mapping possible military domain novelty types to a domain-independent ontology developed as part of a theory of novelty. Characterizing the possible space of novelty mathematically and ontologically will allow us to experiment with agent designs that are coming from the DARPA SAIL-ON program in relevant military environments. Utilizing the same techniques as being used in laboratory experiments, we will be able to measure agent ability to detect, characterize, and accommodate novelty.

## 1 MOTIVATION

The military encounters the concept of novelty constantly. Particularly, when faced against an adversary, they are forced to solve these novelties, often with little decision-making time. For example, in 1999, NATO forces conducted an aerial bombing against Serbia during the Kosovo War, called Operation Allied Force. This proved to be a learning experience for the NATO allies, since multiple, novel events were encountered throughout the operation. First, Serbian troops were more steadfast and better equipped than expected. Second, Serbs relied heavily on surface-to air missiles (SAM) and anti-aircraft artillery (AAA) which required Allied forces to fly higher than anticipated, making target location challenging. Third, the Serbs unexpectedly used missiles with radar emission control. All these novelties needed to be solved effectively by the Allied forces in the moment, while navigating other difficulties such as unruly weather and terrain. (Lambeth, 2001) Being able to define and characterize novelty like these within the military domain is one goal of the DARPA SAIL-ON project.

## 2 BACKGROUND

Creating AI agents to solve novel events needs further study within the field of AI. It is an important problem to solve to ensure that agents are robust enough to transition to the real world where novelty flourishes. There have been a few characterizations of novelty, including an attempt to unify various formalizations and models of novelty. (Boult et al. 2021) The SAIL-ON Novelty Working Group (NWG) has explored the *theory of novelty* and created a novelty hierarchy to classify novelties based on the element of the open world affected. (Senator, 2019) We will use this knowledge to create relevant novelties for the DOD domain.

An important aspect of any command and control (C2) system is the ability to learn new situations over time and discovering the best way to react to them so that the system can remain relevant and adapt to current conditions. Moreover, one of the most important aspects of ML/AI-based mission planning application is the requirement for training data. When possible, reliable and sufficiently large historical data is the best way to create models that will provide the best fidelity for machine learning. However, in military applications the amount and quality of historical data is either not large enough or too unreliable to create high fidelity models. This requires studying simulators to create significant data sets to exercise machine-learning methods. One goal of the SAIL-ON program is to generate not only scenarios, but useful novelties that can be given to AI agents in order to truly test their readiness for the unexpected.

For the warfighter, novelty might be best described as something unexpected or not considered given previous training and intelligence. However, the warfighter has a vast array of training and intelligence for "what if" scenarios, so they are ideally prepared for nearly every reasonable novelty they might encounter. Thus, the concept of novelty is not new to the military and has been implicitly used to prepare for various missions, whether they be tactical, operational, or strategic. In fact, the term Black Swan Events is used by the military to categorize major unexpected events that might have catastrophic consequences. (Burns and Miller, 2014) We will not limit our novelties to just characterizing Black Swan Events, but also more nuanced events to test our agents against. We will use the novelty hierarchy developed by the NWG to categorize novelties within the DOD. Additionally, we will propose a mathematical method to model various types of novelties.

## 3 EFFECTS OF NOVELTY ON AGENTS WITH ARTIFICIAL INTELLIGENCE

Current artificial intelligence (AI) systems excel at narrowly scoped, closed world tasks such as playing board games (Silver, 2017) and enhancing image classification. These AI systems, however, are known to struggle against out-of-distribution inputs (Langely, 2020), and their performance drops severely when they are tested in uncontrolled and unforeseen conditions (Chao, 2020) which are commonly faced by military warfighters. For AI systems and humans to work together in military domains, AI systems need to be able to detect, characterize, and accommodate novel situations in the open world environments where warfighters operate.

To further illustrate this point, we will refer to a simulated tactical scenario called Scenario Zero in which a blue force fighter jet (controlled by an AI agent), which helps make its decision based on various sensor data, is tasked to destroy a red enemy's ammo storage site while evading two nearby red enemy force's surface-to-air missile (SAM) launcher. In a pre-novelty scenario, the SAM's missile range is held constant, and the AI agent can successfully execute its task without having the fighter jet shot down. However, when considering the rapid advancements in technology, it is an obvious oversight to assume that the SAM's missile range will always be held constant. In the post-novelty, open world scenario, the SAM's range is increased beyond what the blue fighter jet was previously accustomed to, causing the agent to be shot down before executing its task.

Rather than replanning its course to avoid being shot down, the AI agent in Scenario Zero does not even recognize that a novel situation has occurred, and it continues to send more assets to carry out its task only for them to be destroyed. In an ideal scenario where AI agents are created to be aware of the possibility of novelty, a more robust agent could instead learn to detect, characterize, and accommodate real world novelties.

## 4 THEORY OF NOVELTY

Philosophically, everything we encounter in the real world is novel. As the Greek Philosopher Heraclitus says: "The only constant is change." However, within an AI system, there are parameters to define an environment, which makes defining novelty for an agent slightly easier, but potentially more unrealistic. For example, if an AI agent is told that in its world model missiles have a max range of 50km, but it encounters a missile that travels 51km, then is that novel? To the agent, assuming it has no way to derive slight variations in parameters, it technically is novel. However, small variations in parameter values are expected in the real world.

*Table 1: Open World Novelty Hierarchy*

| | | | |
|---|---|---|---|
| Single Entities | Phase 1 | 1 | **Objects:** New classes, attributes, or representations of non-volitional entities. |
| | | 2 | **Agents:** New classes, attributes, or representations of volitional entities. |
| | Phase 2 | 3 | **Actions:** New classes, attributes, or representations of external agent behavior. |
| Multiple Entities | | 4 | **Relations:** New classes, attributes, or representations of static properties of the relationships between multiple entities. |
| | | 5 | **Interactions:** New classes, attributes, or representations of dynamic properties of behaviors impacting multiple entities. |
| Complex Phenomena | Phase 3 | 6 | **Rules:** New classes, attributes, or representations of global constraints that impact all entities. |
| | | 7 | **Goals:** New classes, attributes, or representations of external agent objectives. |
| | | 8 | **Events:** New classes, attributes, or representations of series of state changes that are not the direct result of volitional action by an external agent or the SAIL-ON agent. |

Novelty requires a relational understanding - something is not novel in and of itself, instead, it must be novel when compared to something else. There must be experience, ground truth, or memory to compare novelties. (Muhammad et al. 2021) In an AI system, an agent within its environment either has a model of the world or training experience to compare its new observations and be able to detect novelty.

The previous example, with the missile traveling at 51km, the change is technically a novelty, although it most likely would not be impactful to the agent (depending on how the agent is trained) and might be considered a nuisance novelty (Boult et al., 2021). These sorts of novelties are not ones we want to focus on for our study. However, if these types of novelties do occur, then there should be well defined metrics to account for them.

The NWG created an Open World Novelty Hierarchy (Table 1) (Doctor et al. 2022) to classify novelties into 8 different levels. In one sense, each level increases in complexity, however, this does not mean that detection, characterization, or accommodation becomes increasingly difficult as well. Also, some novelties could be classified into more than one level of the hierarchy, but we fit each novelty into only one level that fits best. Although the hierarchy does not necessarily cover every possible novelty, it does capture a rich variety.

# 5 Novelty Within the DOD

Within the DOD, novelty can cover a vast array of scenarios. Decisions within the DOD are often categorized with three levels of warfare - tactical, operational, and strategic. (Doctrine for the Armed Forces of the United States, 2013) We will focus on specific examples given by the tactical Scenario Zero explained in Section 3. Note that the following examples of novelty include those that could potentially make the fighter jet's mission more difficult. However, novelty could include events that make its mission easier, but these are not important novelties on which we want to focus.

## 5.1 Single Entities

In Scenario Zero we classify the SAM and the fighter jet as volitional entities since they both have goals to strike at each other. Levels 1 through 3 of the novelty hierarchy contain novelties that affect only a single entity within the scenario.

### 5.1.1 Objects

Non-volitional objects within a tactical military scenario include entities that either block or impede a blue agent's ability to complete its mission.

For Scenario Zero, the following novelties could take place:

- Addition of nearby no-fire entities, such as civilian towns or embassies.
- Arduous or view-limiting terrain.

### 5.1.2 Agents

Volitional entities would include red or blue agents within our scenario. Novelty within those entities would include increasing the agent's capabilities or characteristics about each of those entities.

For Scenario Zero, the following novelties could take place:

- Advancing the capabilities of one or more SAMs, such as increased missile range, increased speed of warhead, more sophisticated warheads (heat-seeking), increased number of warheads, etc.
- Ammo storage has an enemy defender with the ability to fire weaponry.
- Additional SAMs.

### 5.1.3 Actions

This is where novelties affect the behavior or action space of an external agent, but the behavior itself it not necessarily changed. In a military scenario, the way in which an external agent can achieve its mission is modified or changed.

For Scenario Zero, the following novelties could take place:

- One or more SAMs become mobile.
- One or more of the SAMs becomes a decoy.

## 5.2 Multiple Entities

The next three levels refer to novelties that will impact multiple entities, either between volitional entities, including the SAIL-ON agent, or non-volitional entities.

Relational level novelties are static changes in the relationship between two or more entities, but do not affect their behaviors. Whereas, interactional level novelties are a dynamic change of the behavior between two or more entities. Within a tactical scenario, these levels of novelty could be between blue agents only, red agents only, or both.

### 5.2.1 Relations

For Scenario Zero, the following novelties could take place:

- Changing the location of the SAM to a more difficult terrain or closer to no-fire entities.
- One or more sensors break and transmit no data to the fighter jet.

### 5.2.2 Interactions

For Scenario Zero, the following novelties could take

place:

- Survivability of SAM or ammo storage site increases.
- Making the SAM mobile and actively to move closer to the fighter jet.
- One or more sensors breaks or becomes compromised and conveys false data to the fighter jet.

### 5.3 COMPLEX PHENOMENA

In general, these novelties affect the environment or agents as a whole. This makes implementation often more difficult. Scenario Zero is a simple scenario, so there are not too many meaningful examples of the following novelty levels. However, each novelty level will be explained within the scope of the DOD. Also, a richer scenario would be easily thought to incorporate more examples of novelty here.

#### 5.3.1 Rules

Rules affect all entities within the environment. This is an important military novelty, since planning out every detail of a mission is incredibly difficult. The warfighter might encounter new roadblocks or rule changes for how they can carry out and complete their mission.

For Scenario Zero, the following novelties could take place:

- No-fire zones for all entities.
- Time constraints for when active engagement is permitted.
- Every entity has a time delay when executing an order to fire.

#### 5.3.2 Goals

Goals can affect one or more agents' reason for behaving as they do. The goal of the SAIL-ON agent should not change as that defeats the purpose of the experiment. Within the military scenarios, goals are usually straightforward between blue and red entities.

For Scenario Zero, the following novelties could take place:

- One or more of the SAM sites are no longer tasked with the mission defending the ammo storage site. Instead they are tasked with the mission of destroying the fighter jet.

#### 5.3.3 Events

Events are changes to the environment that are not due to any action by a volitional agent. Events can have a wide definitional scope, since it covers any possible changes outside the volitional entities themselves. However, we want to make sure event novelties are kept reasonable and realistic.

For Scenario Zero, the following novelties could take place:

- Introduction of a red enemy fighter jet in the middle of the scenario.

## 6 MATHEMATICALLY MODELING NOVELTY

To create a rich variety of novelties that model the uncertainty and messiness of the real word, we introduce statistical distributions. Not every novelty will have a proper statistical distribution. For example, in Scenario Zero, if one of the SAMs were changed to be a decoy, that would be a static novelty that does not include any variation. On the other hand, we can add to the complexity of this novelty, by adding more parameters for how a decoy SAM operates. This is simply to say that statistical distributions can be useful to add richness and help model the real world, but are not necessary to introduce novelty.

Novelties often involve parameter changes in continuous variables. For example, in Scenario Zero, missile range, missile speed, red/blue agent survivability, and number of warheads available are all quantitative variables.

We can model these novelties with statistical distributions. For continuous variables, we want to use a Normal distribution or a Uniform distribution. Of course, the benefit of a Uniform distribution is the strict minimums and maximum parameter values, to avoid overlap or potential impossible values. For SAIL-ON specifically, each novelty performers usually further break down novelties into easy, medium and hard. So, there will be three different versions of the same novelty.

Let's use missile range, defined as $R$, as an example. We will use a model with a normal distribution, with a mean $\mu$ and a standard deviation $\sigma$.

$$R \sim N(\mu, \sigma)$$

We can create three distributions, for easy ($R_E$), medium ($R_M$), and hard ($R_H$), defined as:

$$R_E \sim N(50km, 2km)$$
$$R_M \sim N(55km, 2.5km)$$
$$R_H \sim N(60km, 3km)$$

which are graphed in Figure 1. We can sample novel missile ranges within each of these distributions,

depending on the level of difficulty. Further, the three difficulty levels will further help us test the AI agent ability to characterize, detect, and accommodate for different variations of the same novelty. Furthermore, the method in which an agent responds to the three difficulty levels of novelty could be vastly different and provide further information about how the agent handles different situations.

s

*Figure 1: Novelty Distributions for Missile Range, R*

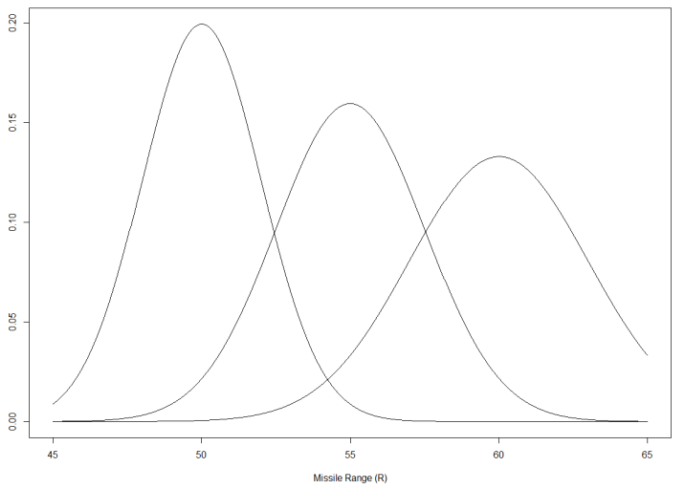

SAIL-ON performers have various approaches to defining novelty. Some might use a distribution as described above, often using uniform distributions or discrete uniform distributions, which help to avoid overlap between the easy, medium, and hard levels. Others might use static numbers for some of their novelties. Using the missile range example, some might simply choose $R_E = 50km$, $R_M = 55km$, and $R_H = 60km$. Currently, we are using static numbers to begin our testing. Next, we plan to move toward a statistical approach to create a richer, more realistic model of novelties in the open world. Note that this statistical distribution approach will require a large set of samples to collect enough data for creating accurate metrics on agent performance.

The missile range example was for a specific continuous, quantitative variable. However, each novelty will have their own distribution assigned to them. Sometimes a normal distribution or uniform distribution will not be the most fitting. For binary variables, such as survivability, we can use the Bernoulli distribution. For rare events, such as sensor failure, we can use the Poisson distribution.

## 7 GENERATION OF NOVELTY BATTLES

Over the last few years, CUBRC has developed the Simulation Orchestration for Learning and Validation Environment (SOLVE), using and integrating several simulators for data generation purposes. SOLVE, as a framework, can leverage video games and military simulators to rapidly test and evaluate mission planning algorithms. SOLVE currently integrates with the Advanced Framework for Simulation, Integration, and Modeling (AFSIM) and One Semi-Automated Forces (OneSAF). AFSIM is currently being leveraged to simulate Multi-Domain Dynamic Targeting (MD2T) scenarios. These scenarios are used to evaluate planning algorithms that synchronize the application of kinetic and non-kinetic effects on targets.

A key innovation under the SAIL ON effort is the use of SOLVE to generate *campaigns* of novelty experiments, which we call *battles*. Campaigns are a set of sequential battles executed on a consistent simulation environment to facilitate evaluation of planning systems to new situations and their ability to react to them. In order to generate all necessary resources for a campaign, as well as attributing purpose and ownership, we create a configuration file in SOLVE which tells the library all the information it needs. The following operations are encoded into the variations:

- Move blue platform X to an exact location
- Move blue platform Y to a uniformly sampled location within a bounding box
- Move red platform Z to a normally distributed location within a bounding box

Each campaign design has inherent variation within the simulation to model real world messiness. Novelties can be considered a type of variation - the only difference being novelties are introduced after a given battle within the campaign design. This design pattern allows for the decoupling of the two different types of variations. A novel object can support multiple variations (i.e., multiple instances of novelties). For example, a novel object can update the maximum missile range assumed for a weapon system to X meters after N battles, simulating the introduction of a new intelligence to the system.

By leveraging SOLVE, CUBRC provides SAIL-ON with a stable, initial, common runtime environment that can support the integration of containerized software capabilities. SOLVE is typically deployed to a Kubernetes computing cluster. As shown in Figure 3 in the Appendix, the architecture provides planners with a simulator-agnostic execution engine that generates novelty battles with the capability to:

- Execute different plans (e.g., branch and sequel graphs) and evaluate its impact to other AI agents and novel situations.

- Validate AI behavior/tasking models against human users.
- Facilitate the integration of planners at different command levels with decision aids.
- Access a mission dashboard which provides metrics and assessments of single runs or an aggregation of battles.
- Provide a data strategy that persists and retrieves plan and analytics data for evaluation or dissemination. Figure 4 in the Appendix shows SOLVE's UI that tracks and accesses running instances or historical runs.
- Create a diverse set of Design of Experiments (DOE) with auto variant exploration and novelty generation.

### 7.1 METRICS FOR AGENT EVALUATION

We need metrics that are used for any agent in the open world, but also scenario specific metrics. Table 2 shows our current agent evaluation metrics for Scenario Zero.

*Table 2: Metrics for Scenario Zero*

|  | Mission Status | | |
|---|---|---|---|
| **Metric** | *Fail* | *Abort* | *Win* |
| SAIL-ON Agent Score | 0 | .5 | 1 |
| Number of Targets Prosecuted | 0 | 0 | 1 |
| Number of Friendly Casualties | 1+ | 0 | 0 |
| Were Time Limits Exceeded | Yes | No | No |
| Novelty Accurately Detected | No | Yes | Yes |
| Novelty Accurately Characterized | No | Yes | Yes |

There are three options for the SAIL-ON Agent Score. The agent receives a 0, when the target is not destroyed and there are more than one friendly casualties, or the scenario time limit is exceeded. The agent receives a score of 0.5 when the target is not destroyed and there are no friendly casualties within that scenario time limit. A score of 0.5 indicates that novelty was detected accurately, but was only characterized sufficiently to justify aborting the mission. The agent receives a score of 1 when the target is destroyed and there are zero friendly casualties within the scenario time limit. A score of 1 indicates that the novelty was detected accurately and was characterized successfully.

Next, we will use existing metrics which were created for any open world learning (OWL) environments. In particular we will implement percentage of correctly detected trials (CDT) and average number of instances to detect novelty (IDN) to measure novelty detection, and novelty reaction performance (NRP), asymptotic novelty reaction performance (ANRP), and double-ended novelty reaction performance. (Pinot et al. 2022)

### 8 FUTURE WORK

To create a more interesting experiment, we are working towards creating a more complex scenario which will allow for a wider range novelty. We are designing this scenario to be a realistic scenario for the warfighter, but within the boundaries of SOLVE.

With this new, complex scenario we can create a full list of novelties with corresponding statistical distributions to more fully test the AI agents. The metrics we use will involve both scenario specific mission objectives and metrics proposed by Pinto et al. (2022) for OWL environments.

Currently, we are using a simulated tactical scenario, where the AI agent completely controls the blue fighter jet. However, we envision future use of this research can include the utilization in unmanned systems, or having the AI aid the warfighter within the kill-web.

### 9 ACKNOWLEDGEMENTS

(anthony.hoogs@kitware.com), and Tom Dietterich (tgd@oregonstate.edu), Marshall Brinn (marshall.brinn@raytheon.com), and Jivko Sinapov (jivko.sinapov@tufts.edu).

## 10 APPENDIX

*Figure 3: Sample Solve Simulation Control*

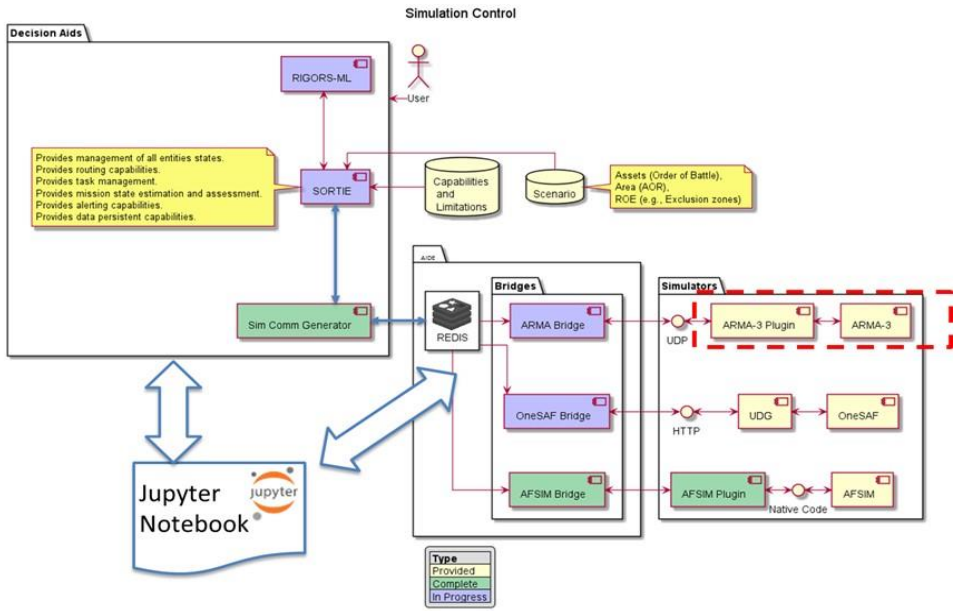

*Figure 4: Current runs monitoring in SOLVE UI*